\documentclass{article}
\usepackage{times,uweetr}
\usepackage{graphics} 
\usepackage{graphicx}
\usepackage{amsmath} 
\usepackage{amssymb}  
\usepackage[bookmarks=true]{hyperref}
\usepackage{enumerate}

\title{To Make a Robot Secure: \\
An Experimental Analysis of Cyber Security Threats Against Teleoperated Surgical Robotics}

\author{T. Bonaci$^{1}$, J. Herron $^{1}$, T. Yusuf$^{2}$, J. Yan$^{1}$, T. Kohno$^{2}$, H. J. Chizeck$^{1}$
\thanks{This work is supported by the National Science Foundation, Grant \# CNS-1329751.}
\thanks{$^{1}$ Department of Electrical Engineering, University of Washington,
       	185 Stevens Way, Seattle, WA, 98195, USA
        {\tt\small \{tbonaci, jeffherr, junjiey, chizeck\}@uw.edu}}%
\thanks{$^{2}${Department of Computer Science and Engineering, University of Washington, 
		185 Stevens Way, Seattle, WA, 98195, USA}
        {\tt\small \{yusuft, yoshi\}@cs.washington.edu}}%
\\
$^{1}$ Department of Electrical Engineering, University of Washington \\
$^{2}$ Department of Computer Science and Engineering, University of Washington \\
Seattle WA, 98195-2500\\
}

\reportmonth{April}
\reportyear{2015}
\reportnumber{0002}

\title{\LARGE \bf
To Make a Robot Secure: \\
An Experimental Analysis of Cyber Security Threats Against Teleoperated Surgical Robotics\footnote{This work is supported by the National Science Foundation, Grant \# CNS-1329751.}}

\author{Tamara Bonaci$^{\dagger}$, Jeffrey Herron $^{\dagger}$, Tariq Yusuf$^{\ddagger}$, Junjie Yan$^{\dagger}$, Tadayoshi Kohno$^{\ddagger}$, Howard Jay Chizeck$^{\dagger}$
\thanks{Department of Electrical Engineering, University of Washington,
       	185 Stevens Way, Seattle, WA, 98195, USA
        {\tt\small \{tbonaci, jeffherr, junjiey, chizeck\}@uw.edu}}%
\thanks {{Department of Computer Science and Engineering, University of Washington, 
		185 Stevens Way, Seattle, WA, 98195, USA}
        {\tt\small \{yusuft, yoshi\}@cs.washington.edu}}%
        }
        
\begin{document}

\maketitle

\begin{abstract}
Teleoperated robots are playing an increasingly important role in military actions and medical services. 
In the future, remotely operated surgical robots will likely be used in more scenarios such as battlefields and emergency response. But rapidly growing applications of teleoperated surgery raise the question; what if the computer systems for these robots are attacked, taken over and even turned into weapons?

Our work seeks to answer this question by systematically analyzing possible cyber security attacks against Raven II\textsuperscript{\textregistered}, an advanced teleoperated robotic surgery system. We identify a slew of possible cyber security threats, and experimentally evaluate their scopes and impacts. We demonstrate the ability to maliciously control a wide range of robots functions, and even to completely ignore or override command inputs from the surgeon. We further find that it is possible to abuse the robot's existing emergency stop (E-stop) mechanism to execute efficient (single packet) attacks.

We then consider steps to mitigate these identified attacks, and experimentally evaluate the feasibility of applying the existing security solutions against these threats. The broader goal of our paper, however, is to raise awareness and increase understanding of these emerging threats. We anticipate that the majority of attacks against telerobotic surgery will also be relevant to other teleoperated robotic and co-robotic systems. 

\end{abstract}

\vspace{-5pt}
\section{Introduction}\label{sec:Introduction}
Recent developments in robotics have had a profound impact in medicine. There has been a 20\% yearly increase rate in the number of medical robots sold~\cite{Medical_robots_sales}. The use of robots in medical procedures has been shown to result in better surgical outcomes and faster recovery, thus enhancing the delivery of medical services to the patients~\cite{hu2009comparative}.  

The research and development of teleoperated robotic surgery systems is recognized as the next step in medical robotics~\cite{rosen2006doc}. Teleoperated surgical robots will be expected to use a combination of existing publicly available networks and temporary {\it ad-hoc} wireless and satellite networks to send video, audio and other sensory information between surgeons and remote robots~\cite{lum2007field}. It is envisioned these systems will be used to provide immediate medical relief in under-developed rural terrains, areas of natural and human-caused disasters, and in battlefield scenarios~\cite{harnett2008evaluation}. But what if these robotic systems are attacked and compromised? Recent examples, such as Stuxnet worm, specifically designed to target programmable logic controllers, and blamed for ruining a significant part of Iran's nuclear centrifuges~\cite{falliere2011w32}, exemplify possible issues when a cyber-physical system is targeted explicitly. 

To date, security has not been a concern for telereobotic surgery. Yet researchers have recognized that the open and uncontrollable nature of the communication medium opens these systems to a variety of possible cyber security vulnerabilities~\cite{lee2012cyberphysical}. While a few approaches, focusing mainly on private communication~\cite{tozal2013adaptive} and the ability to verify the robot's side code~\cite{coble2010secure}, have recently been proposed, there is currently little understanding of what the actual risks are.

This lack of understanding of the actual risks is a function of two factors. At the moment, it is not known: (1) how easy it would be for an attacker to compromise a teleoperated surgery system, and (2) what the applications of such a cyber security attack might be. Not being able to answer these questions makes it hard to understand what the challenges to improving cyber security of telerobotic surgery are, much less to address them.  

In this paper, we seek to answer the above questions through an empirical analysis of the robotic surgery platform, {\it Raven II}\textsuperscript{\textregistered}. Our work is experimental, along the lines of much past work that explored the security and privacy properties of emerging technologies, including modern automobiles~\cite{koscher2010experimental,checkoway2011comprehensive} and medical devices~\cite{gollakota2011they,halperin2008pacemakers}.

We seek to provide an informed understanding of risks and defenses, based on an evaluation of real technologies. We make the following specific contributions:

{\bf Attack identification and characterization:} We identify possible cyber security attacks against teleoperated robotic surgery, and classify them, based on the impact they have on a surgeon, into three clasees: intention manipulation, intention modification and hijacking attacks. For each of these classes, we characterize the attack surface exposed in teleoperated robotic surgery.

{\bf Vulnerability analysis:}
For each characterized attack surface, we thoroughly analyze one or more practical examples, and assess the level of the actual impact on the surgical procedure. We demonstrate the ability to maliciously control a wide range of robots functions, and even to completely ignore or override surgeon's inputs. We further find that it is possible to abuse the robot's emergency stop (E-stop)  to execute efficient (one packet) denial-of-service attacks.

{\bf Risk assessment and defense directions:}
For the uncovered vulnerabilities, we consider the question of the cost and the benefit for an attacker. Our results, unfortunately, show that an attacker can easily and quite efficiently disrupt a surgical procedure. We thus propose a few simple first steps towards making telerobotic surgery resilient to some of the identified attacks. We experimentally investigate the feasibility of the proposed security solutions to telerobotic surgery, and find that they can easily be applied without any significant impact on a system's usability and performance. Finally, we discuss some possible legal implications that the identified attacks may impose on a surgeon, a hospital, and manufacturer.

{\bf Challenges specific to teleoperated procedures:} During our experimental analysis, we observe several cyber security challenges specific to teleoperated surgery (and other teleoperated robotic systems). The most surprising among them may be the tension between the role of the E-stop feature in the benign scenario and under attack. While under normal circumstances, the existence of E-stop increases the safety of patients, nearby human operators, and the robot, the same E-stop feature may be abused by an attacker to decrease the safety and security of patients, operators and robots.

\vspace{-5pt}
\section{Related Work}\label{sec:Related_work}

\subsection{Raven II\textsuperscript{\textregistered} Surgical Robot}\label{subsec:Raven}
The {\it Raven II} is a teleoperated robotic system designed to support research in advanced techniques of robotic-assisted surgery~\cite{rosen2006doc,hannaford2013raven}. It is the first experimental platform in surgical robotics capable of supporting both software development,  experimental testing, and medical (surgical) training. It is commercially available from Applied Dexterity~\cite{Applied_dexterity}, and it is currently a research platform at 12 universities across the U.S., Canada, France and United Kingdom. 

\begin{figure}[t!]
\vspace{-50pt}
\begin{center}
\includegraphics[width=2.5in]{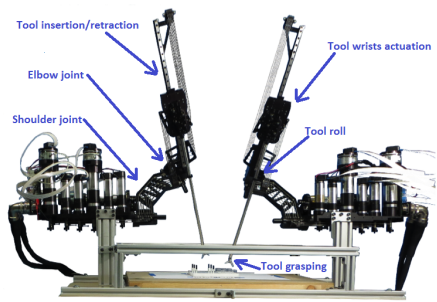} 
\end{center}
\vspace{-70pt}
\caption{The Raven II system consists of two 7-degrees-of-freedom surgical manipulators. The motion axes of the robot are: shoulder joint, elbow joint, tool insertion/retraction, tool roll, tool grasping, tool wrist 1 actuation and tool 2 wrist actuation.}
\label{fig:Raven_labeled}
\vspace{-15pt}
\end{figure}

As depicted in Figure\ref{fig:Raven_labeled}, the {\it Raven II} consists of two 7-degrees-of-freedom (DOF) surgical manipulators, divided into three main subsystems: the static base that holds all seven actuators, the spherical mechanism that positions the tool, and the tool interface. The motion axes of the robot are: {\it shoulder joint} (rotational), {\it elbow joint} (rotational), {\it tool insertion/retraction} (linear), {\it tool roll} (rotational), {\it tool grasping} (rotational), {\it tool wrist 1 actuation} (rotational), and {\it tool wrist 2 actuation} (rotational). DC motors mounted to the base actuate all motion axes, and the motors of the first three axes have power-off brakes to prevent tool motion in the event of a power failure.

The {\it Raven II} software is based on open standards, including Linux and Robot Operating System (ROS)~\cite{quigley2009ros}. The low-level control system includes real-time Linux processes (modified by the RT-Prempt Config kernel patch), running at a deterministic rate of 1000 Hz. Key functions running inside the 1000 Hz servo-loop are: (i) coordinate transformations, (ii) forward and inverse kinematics, (iii) gravity compensation, and (iv) joint-level closed-loop feedback control. The link between the control software and the motor controllers is a USB 2.0 interface board, designed with eight channels of high-resolution 16-bit digital-to-analog conversion for control signal output to each joint controller, and eight 24-bit quadrature encoder readers. The board can perform a read/write cycle for all 8 channels in 125 microseconds. The two {\it Raven II} arms are controlled by a single PC with two USB 2.0 boards.

In the {\it Raven II} system, surgeon control inputs are collected through a surgical control console. Control inputs and robot feedback, which include video and haptic information, are transmitted using the communication standard for surgical teleoperation, Interoperable Telesurgery Protocol (ITP)~\cite{king2009preliminary}. ITP allows communication between heterogeneous surgical consoles (masters) and manipulators (slaves), regardless of their individual hardware and software. 

In recent years, the {\it Raven II} has been evaluated in several extreme environments scenarios~\cite{lum2007field,harnett2008evaluation}. In the HAPs/MRT field experiment~\cite{lum2007field}, the {\it Raven II} robot was deployed in the Mojave desert. It was controlled across the internet, with the final link being a UAV-enabled wireless network. In that experiment, the following network states were recognized as critical for reliable performance~\cite{lum2009effect}: (i) communication latency, (ii) jitters, (iii) packet delays, out-of-order arrivals and losses, and (iv) devices failures. 

In addition to these stochastic but benign network patterns, surgeon-robot communication over publicly available
networks expose telerobotic surgery to problems most likely not present in hospital settings. Due to the open and uncontrollable nature of communication networks, it becomes easy for malicious entities to jam, disrupt, or take over the communication between a robot and a surgeon. 

\subsection{Security of Cyber-Physical Systems}\label{subsec:CPS_security}
In recent years, several research initiatives have begun to address security and privacy issues of a variety of cyber-physical systems (CPS). We give a brief overview of the state-of-the-art for CPS classes closely related to telerobotic surgery: networked control, medical and robotic surgery.

\subsubsection{Security of Networked Control Systems}
Researchers recently showed that attacks against networked control systems and wireless sensor networks can be mitigated by relying on the system's dynamics (see, e.g.,~\cite{amin2009safe,cardenas2008research,mo2009secure}). In~\cite{mo2009secure}, for example, authors assumed that the system's dynamics are linear, and showed that a simple optimal controller and a Kalman filter can be used to guarantee the desired probability of detecting attacks, such as replay, false data injection and integrity attacks, under a certain model. Their proposed approach is based on trading off the cost assumption with the probability of attack detection. 

\subsubsection{Security of Medical Systems}
The importance of privacy and security for tele-medical applications was first recognized in the mid-1990s, in~\cite{makris1997network,wozak2007end}. After the establishment of the Health Insurance Portability and Accountability Act (HIPAA)~\cite{HIPAA}, patients privacy became for a while the primary concern, and researchers typically focused on protecting the confidentiality of transmitted and stored patient data. For example, in~\cite{yang2003secure}, the authors consider security issues related to the medical data in multimedia form. They present a simulated surgery procedure, and introduce an idea of a smart surgery room, with monitoring actions of participating medical personnel.

More recently, researchers recognized that many modern implantable medical devices, including pacemakers and implantable
cardioverter-defibrillators, are vulnerable to a variety of attacks, allowing attackers to wirelessly obtain private patient information and change device settings in ways that can directly impact patient health~\cite{gollakota2011they,halperin2008pacemakers}.

\subsubsection{Security of Telereobotic Surgery}
Very recently, motivated by the {\it Raven II} extreme operation experiments~\cite{harnett2008evaluation,lum2007field}, researchers recognized importance of cyber security for telerobotic surgery~\cite{lee2012cyberphysical,tozal2013adaptive,coble2010secure}. In~\cite{coble2010secure},  authors developed a light-weight software tool to verify the robot's side code. In~\cite{tozal2013adaptive}, authors developed an information coding approach to protect communication privacy and reliability. In~\cite{lee2012cyberphysical}, the use of the Transport Layer Security (TLS) protocol was proposed to ensure confidentiality, authentication and authorization of the ITP.
\vspace{-5pt}
\section{Vulnerability Analysis}\label{sec:Vulnerability_analysis}

\subsection{Attacker Model}\label{subsec:Attacker_model}
\begin{figure}[h!]
\vspace{-70pt}
\begin{center}
\includegraphics[width=3in]{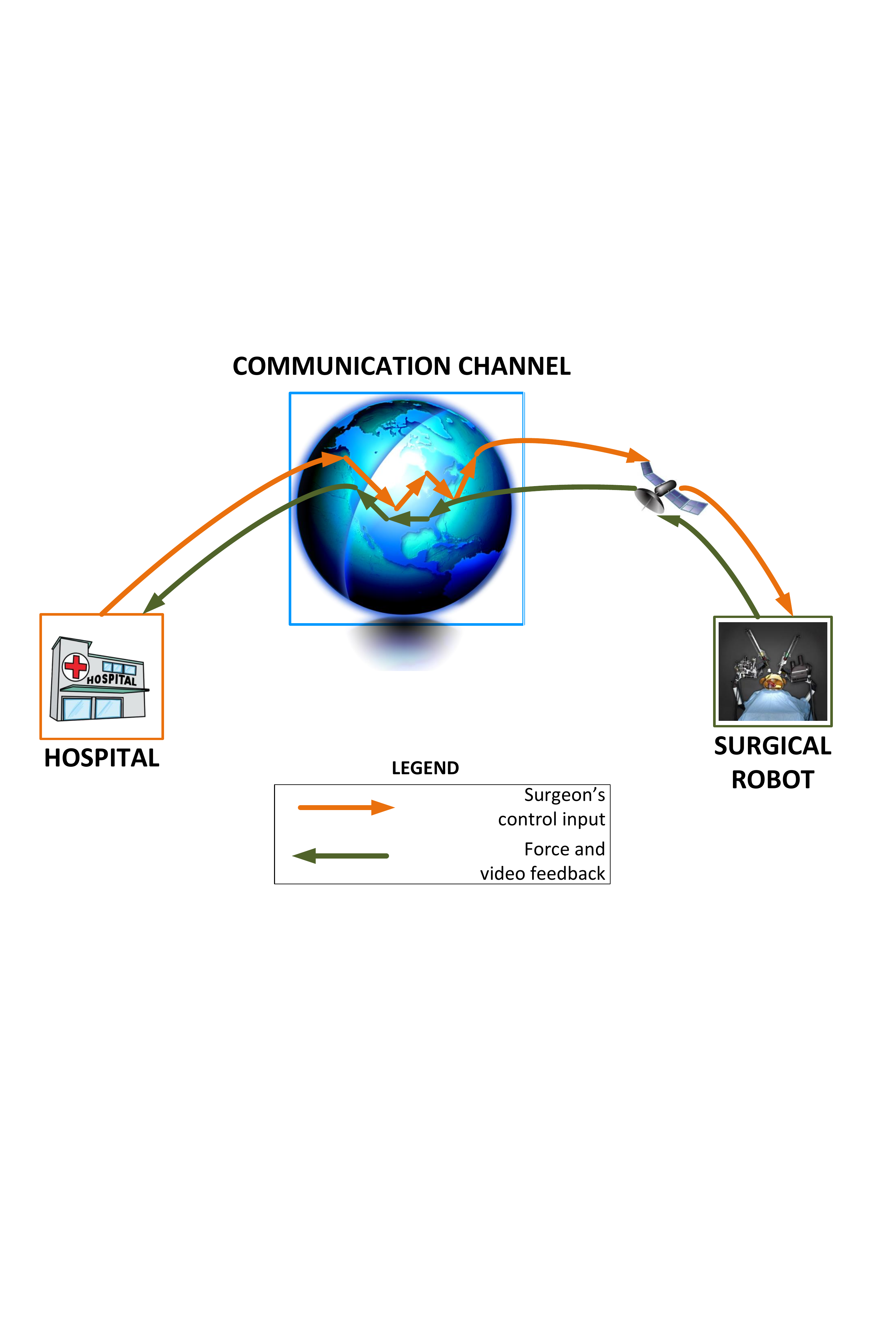} 
\end{center}
\vspace{-110pt}
\caption{Visualization of a typical telerobotic surgery setup. Dashed lines indicate wireless links, solid lines pre-established network connections (either wired or wireless). Orange color indicates surgeon's control messages, and green color robot's feedback messages.}
\label{fig:Raven_setup}
\vspace{-5pt}
\end{figure}

Telerobotic surgery is envisioned to be used in extreme conditions, where robots will have to operate in low-power and harsh conditions, with potentially lossy connection to the internet. The last communication link may potentially even be a wireless link to a drone or a satellite, providing connection to a trusted facility (potentially a large hospital with an established infrastructure), as depicted in Figure \ref{fig:Raven_setup}.

In such operating conditions, we recognize two attack vectors are feasible: (1) {\it endpoint compromise}, where either a surgeon's control console or a robot can be compromised, and (2) {\it network and communication-based attacks}, where an attacker may intercept the existing network traffic, inject new malicious traffic, or both. 

Endpoint compromises are less interesting since physical access to either side will likely be strictly monitored. Network and communication-based attacks thus represent a more feasible way to compromise the system. Moreover, due to their abundance and variability, mitigating these attacks is likely to be intellectually challenging, making this the most difficult part of the system to protect. The most likely point of attack appears to be between the network uplink and a surgical robot. Since communication will likely be wireless, on-the-field attackers will be able to disrupt the link or manipulate traffic contents. In the rest of the paper, we thus focus on disruption and manipulation attacks against surgeon-robot communication links.

\vspace{-5pt}
\subsection{Attack Classification}\label{subsec:Attack_classification}
Based on their impact on surgeons, we categorize possible attacks into three categories: (a) {\it intention modification}, (b) {\it intention manipulation}, and (c) {\it hijacking} attacks. 

{\bf Intention modification} attacks occur when an attacker directly impacts a surgeon's intended actions by modifying his/her messages while packets are in-flight, and a surgeon has no control over them. These attacks are relatively easy to observe when executed correctly, through e.g., unusual robot movements, robot becoming randomly engaged or disengaged, or unusual delays in movements.

{\bf Intention manipulation} attacks occur when an attacker only modifies feedback messages (e.g., video feed, haptic feedback), originating from a robot. A surgeon's messages (and his/her intent) are assumed to be valid. These attacks can prove to be more difficult to mount, simply because of the amount of data that a robot transmits, but if executed correctly, these attacks may be harder to detect and prevent, since they are quite subtle. As the feedback is assumed to be valid, a surgeon's (valid) actions may unintentionally end up becoming harmful to a patient.

In {\bf hijacking} attacks, a malicious entity causes the robot to completely ignore the intentions of a surgeon, and to instead perform some other, potentially harmful actions. Some possible attacks includes both temporary and permanent takeovers of the robot, and depending on the actions executed by the robot after being hijacked, these attacks can be either very discreet or very noticeable.

In addition, we consider the role an attacker needs to assume within the system in order to be able to mount an attack, and with respect to that, we classify attackers into two groups: (1) {\it network observer}, and (2) {\it network intermediary}. 

A {\bf network observer} initially eavesdrops on information exchange between a surgeon and a robot, and based on the collected information, starts inserting false messages into the network, while still allowing both the benign parties to communicate directly.

A {\bf network intermediary} (i.e., a {\it man-in-the-middle} (MitM) attacker) assumes a role of an intermediary between a robot and a surgeon, thus completely preventing the benign parties from communicating directly. In a real life attack scenarios, this can  be done using methods such as ARP poisoning~\cite{abad2007analysis}.

\vspace{-5pt}
\section{Experimental Analysis}\label{sec:Experimental_analysis}
\subsection{Experimental Setup}
\begin{figure}[h!]
\begin{center}
\vspace{-40pt}
\includegraphics[width=2.5in]{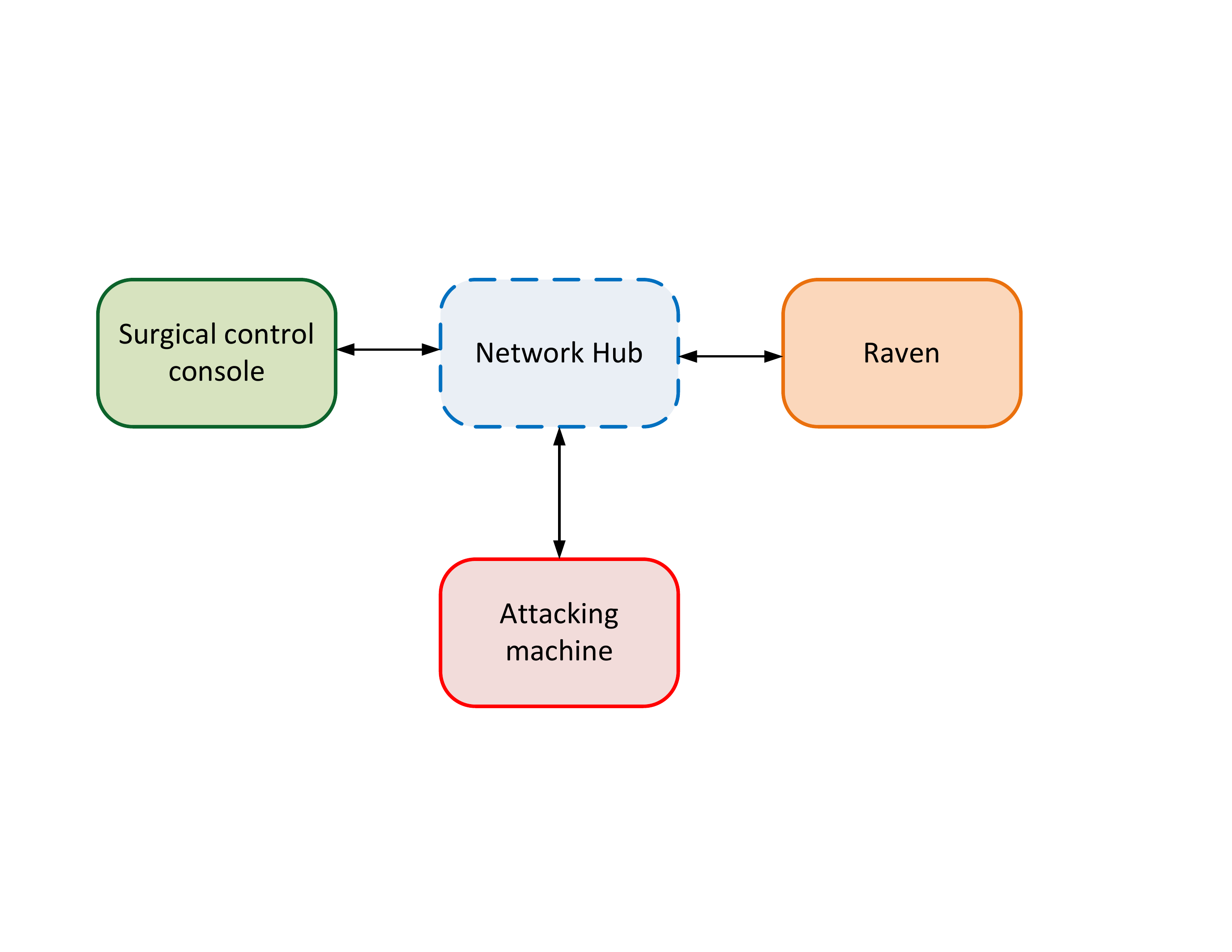} 
\end{center}
\vspace{-50pt}
\caption{Experimental setup: the attacking machine is running either Kali Linux or Windows 7 SP3, with attack implementations written in C\# or Python with the Scapy framework.}
\label{fig:Exp_setup}
\vspace{-10pt}
\end{figure}

To experimentally investigate possible attacks against teleoperated surgery, we establish communication between the surgical control console and the {\it Raven II} robot through a network hub, as depicted in Figure \ref{fig:Exp_setup}. 

This allows us to connect an external computer to the same subnetwork, and use it to observe and modify communication between a surgeon and a robot. Our attacking computer is running either Kali Linux, or Windows 7 SP3, and all of the analyzed attacks are implemented either in C\#, or in Python with the Scapy framework~\cite{biondi2010scapy}.

\vspace{-5pt}
\subsection{Experiment Description}
Our analysis is based on the data collected from experiments involving twenty human participants. This study was approved by the University of Washington Institutional Review Board approval (\#46946 - EB), and all of our subjects were undergraduate and graduate Electrical Engineering and Computer Science and Engineering students, ranging in age from 19 to 28 years. 
We acknowledge that an engineering student's behavior may differ from a surgeon's behavior, but that is acceptable (and an established experimentation method in surgical robotics) since it has been shown in~\cite{lum2009effect}, that both surgical and non-surgical subjects, upon gaining proficiency, achieve similar results in simple surgical robotic tasks, such as the Fundamentals of Laporoscopic Surgery (FLS) block transfer task. 

The {\it FLS block transfer task} is a standard test used to train and test surgeons~\cite{lum2008objective}, where a subject uses robot's graspers to move six rubber blocks, one at the time, from the left side of the FLS pegborad to the right, and then back to the left side. When moving from left to right, a block is picked up from the peg with the left hand, transferred in the air to the right hand, and then placed on the right peg. Hands are reversed when moving from right to left. One trial consists of moving all blocks from left to right and then back from right to left, totaling in twelve transfers. 

\begin{figure}[h!]
\vspace{-30pt}
\begin{center}
\includegraphics[width=1.5in]{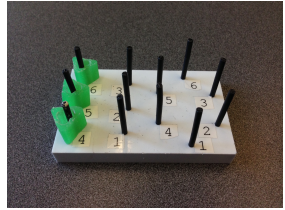} 
\end{center}
\vspace{-40pt}
\caption{A pegboard used in experiments. Each participant was asked to move a rubber block from one of the positions 4, 5, 6 on the left-hand side to one of the positions 2, 3, 6 on the right-hand side.}
\vspace{-10pt}
\label{fig:FLS}
\end{figure}

Due to the nature of our investigation, where we focus on the impact of attacks, rather than on subjects' proficiency, we make three simplifications to the FLS block transfer task:
\begin{itemize} 
\item Instead of six rubber blocks, we only use three.
\item The subjects are asked to move pegs only from left to right, and the right-to-left movement is not needed. 
\item The subjects are allowed to pick up blocks with a grasper of their choosing, and they are not required to transfer blocks in the air from one hand in the other.
\end{itemize}
With these simplifications, our trial consists of moving all blocks {\it only} from left to right, totaling in three transfers. The pegborad we use is depicted in Figure \ref{fig:FLS}.

\vspace{-10pt}

\vspace{-5pt}
\subsection{Intent Modification Attacks}\label{subsec:Modification}
We consider four subgroups of intent modification attacks: (i) reordering, (ii) packet loss, (iii) packet delay and (iv) content modification. In all of these attacks, an attacker is assuming a role of network intermediary (man-in-the-middle), and we achieve that by setting up a direct connection between the surgical control console and the attacking machine, {\it Eve}, which then forwards all the surgeon's messages to the {\it Raven}. 
For all attacks, except packet delay, {\it Eve} is running a simple UDP proxy written in Python with packet interpretation and modification done using \textit{Scapy}~\cite{biondi2010scapy}.

\subsubsection{\bf Surgeon's Intent Reordering}
Intent reordering is a simple zero-knowledge attack where, instead of forwarding a surgeon's packet to the {\it Raven}, we add it to a queue on {\it Eve}, that pops items out in a random order once it reaches the maximum length. As a result, all surgeon's messages are delivered to the {\it Raven} with a negligible delay (caused by the time spent in the queue). The {\it Raven}, however, does not implement all of the received messages. It skips the messages with sequence numbers received out of order, and the effect of skipped messages is a jerky motion of robot's arms, immediately observable by experiment participants.

\subsubsection{\bf Surgeon's Intent Loss}
Intent loss is another zero-knowledge attack, where we randomly drop individual surgeon's packets or groups of packets. As a result of packets being dropped, {\it Raven}'s motion becomes delayed and jerky. We investigate what are the largest tolerable {\it packets dropping rates}, $\eta$, that still result in a reasonably compliant robot. To do so, we wrote a Python script that sweeps the space of allowed $\eta, 0 \leq \eta \leq 1$.

For individual packet drops, we observe that $\eta \geq 0.55$ generally makes the robot operable, but difficult to use, because grasping becomes challenging. When $\eta$ increases to 0.9, the robot becomes almost unusable, in particular when the required movements are small and precise. For group packet drops, we consider groups consisting of 100 packets (10\% of packets transmitted every second), and we find that packet dropping rate $\eta \leq 0.2$ results in a generally operable robot, but the robot's movements are still jerky.

\footnotesize
\begin{table}[ht]
\centering
\begin{tabular}{l c c c c c c c}
\hline
 & Uni, 100-500 & Uni, 400-600 & Uni, 300-700 & No delay & Gauss, 0-200 & Gauss, 100 & Gauss, 100-250\\ [0.5ex] 
\hline\hline
Subj. A & 361 & 248 & 632 & 40 & 118 & 100 & 520 \\
\hline
Subj. B & 154 & 620 & 745 & 33 & 72 & 100 & 87 \\
\hline
Subj. C & 350 & 261 & 229 & 65 & 292 & 137 & 555 \\
\hline
\end{tabular}
\caption{{\footnotesize Time (in seconds) needed to finish one simplified FLS block transfer task under different types of surgeon intent delays. `Uni' denotes uniformly and `Gauss' normally distributed delay. Numbers x-x denote the delay range in ms.}}
\vspace{-10pt}
\label{table:Delay_results}
\end{table}
\normalsize

\subsubsection{\bf Surgeon's Intent Delay}
To mount a surgeon's intent delay attack, we developed our own C\# toolbox that allowed us to control different characteristics of the per-packet delay, including the amount of delay imposed on an individual packet and a type of the delay. We modeled a delay as a random variable, where in every FLS block transfer task, we assumed delays are independent, identically distributed (iid) random variables, and a delay (a single random variable) is added to every surgeon's command and robot's feedback messages. We consider three types of delay: (a) constant, (b) uniformly distributed, and (c) normally (Gaussian) distributed, resulting in ten different simplified FLS block transfer tasks: (1) no delay, (2) constant delay of 300ms, (3) constant delay of 500ms, (4) uniform delay between 100-500ms, (5) uniform delay between 400-600ms, (6) uniform delay between 300-700ms, (7) no delay, (8) Gaussian delay between 0-200ms, (9) Gaussian delay of 100ms, and (10) Gaussian delay between 100-250ms. For each of these tasks, we measure the time needed to complete the task, and we record subjects' assessment of the task difficulty. 

{\it Time Needed to Complete the Task:} Table \ref{table:Delay_results} presents time needed to complete one simplified version of FLS block transfer task under seven different cases of surgeon intent delay attacks (tasks (4)--(10)). Comparing the case of no delay (column 4) with all of the delay cases, we observe that, on average,  delay attack increases the time to complete the task by 5 times. Further, based on these preliminary results, Gaussian delay with delay ranging between 100-250ms seems to have the most severe impact on subjects. 

{\it Subjective Assessment of Task Difficulties:} For each of the simplified FLS tasks (1)--(10), we further ask subjects to evaluate the difficulties of: (i) reaching each of the three blocks, (ii) grabbing the blocks, (iii) moving between the pick-up and put-down locations and (iv) performing the task as a whole, where the allowed difficulties ranged from 0 (easy) to 7 (hard). We collect these self-reported results from 10 subjects.

In order to evaluate the collected results, for each delay attack and each subject, we sum up the four reported difficulty categories, thus obtaining a single number as a representation of the perceived difficulty of an attack. We refer to this number as the {\it delay difficulty score}. We evaluate if there exists a statistically significant difference between delay difficulty scores for different types of attacks. In doing so, we apply the Wilcoxon signed-rank test, since the sample size was small (10 subjects) and the obtained delay difficulty scores were not normally distributed.

Not surprisingly, the cases with no delay are statistically significantly different (largest  P value for trial 1: $p_{1} = 2.34 \cdot 10^{-2}$; largest P value for trial 7: $p_{7} = 3.91^{-3}$.) With the given sample size, however, we were not able to distinguish between the difficulties of different types of delay attacks. Yet, the obtained delay difficulty scores indicate that the Gaussian delay with delay ranging between 100-250ms was perceived to be the hardest by the majority of subjects (reported delay difficulties per subjects: 0, 6, 25, 20, 28, 10, 25, 27, 19, 25, with the first result clearly an outlier).

\subsubsection{\bf Surgeon's Intent Modification}
Leveraging knowledge about the structure of surgeon's packets~\cite{king2009preliminary}, we next modify surgeon's packets on-the-fly before forwarding them to the {\it Raven} through our proxy, {\it Eve}. Some of the attacks we consider:
\begin{itemize} 
	\item Changing the commanded changes in position,
	\item Changing the commanded changes in rotation,
	\item Inverting the grasping states of robotic arms,
	\item Inverting a combination of the above attacks to fully invert left and right robotic arm, and
	\item Randomly scaling the commanded changes in position and rotation.
\end{itemize}
Most of these attacks had a noticeable impact on the {\it Raven} immediately upon launch. In particular, if an attack involved any changes to grasping state, even a modification of a single packet had a profound impact on the FLS block transfer task. Unsurprisingly, the least noticeable case was the attack affecting the positions of robot's arms, as long as the modified changes are within the allowed region. Once the modified changes required too large or too fast changes in the positions of robotic arms, thus effectively requiring too high currents, the robot's safety mechanism clipped the currents, resulting in a noticeable slower robot motion.

\begin{figure}[h!]
\vspace{-30pt}
\begin{center}
\includegraphics[width=4in]{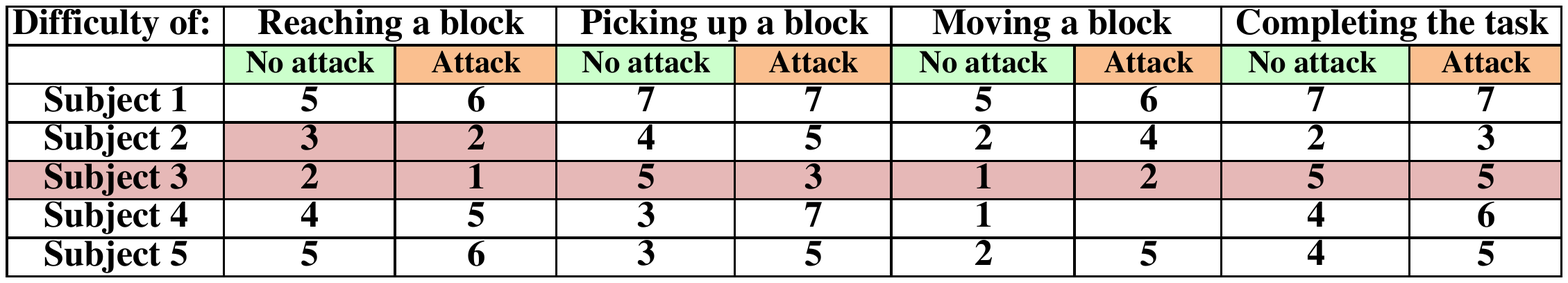} 
\end{center}
\vspace{-170pt}
\caption{Subjective assessment results for surgeons' intent modification attack. The difficulties ranged from 0 (easy) to 7 (hard).}
\vspace{-5pt}
\label{fig:intent_modification}
\end{figure}

We next analyze subject responses to these attacks. As in the delay attack, subjects were aware that attacks might be mounted while they are executing the task, but they did not know when they were being targeted nor which attack was being mounted against them. Detailed results, collected from five subjects, are depicted in Figure \ref{fig:intent_modification}. 

Subjects were asked to evaluate the difficulties of: (i) reaching each of the three blocks, (ii) grabbing the blocks, (iii) moving between the pick-up and put-down locations and (iv) performing the task as a whole, and difficulties ranged from 0 (easy) to 7 (hard). The case of Subject 3 is quite peculiar, since it seems that subject did not notice the attack. Based on the reported results, we conclude that other subjects did notice attacks occurring. However, all of them  carried the task to completion. Moreover, they were able to adjust to the attacks within 1-1.5 seconds time period (even in the case of complete inversions of robot's arms). Performing a random combination of the described attacks did, however, result in several typical errors, such as dropping the block, moving the robotic arms outside of the allowed workspace, or triggering E-stop, in order to avoid undesirable robot's movements.

\vspace{-5pt}
\subsection{Hijacking Attacks}\label{subsec:Hijacking}
In hijacking attacks, an attacker assumes a role of a network observer, who can eavesdrop on packets between a surgeon and a robot, without modifying them. After sufficient reconnaissance, an attacker then inject new, malicious packets into the network, in order to impact the surgical procedure. In our case, the reconnaissance phase required only capturing the current packet sequence number, at which point we were able to take over control of the robot. We consider several subgroups of hijacking attacks, namely (1) sequence number leading attack and (2) force reset.

\noindent{\bf Sequence Number Leading:}
Leveraging again prior knowledge about the structure of surgeon's packet~\cite{king2009preliminary}, we conduct the following sequence number leading attack: we first read a single surgeon's packet, and extracted the current sequence number, $seqNum$, from it. We then added a random offset, $rand$, to the sequence number, $\tilde{seqNum} = seqNum + rand$, 
where the only requirement is that the offset needs to be less than 1000. We composed a new (empty) surgeon packet with the new sequence number, $\tilde{seqNum}$ and sent it to the {\it Raven}. At that point, we took over the control of the robot, since the robot attributed the large jump in sequence numbers to packet drops, and as long as the system did not lose more than a second of data, the operation continued. From this point on (until the sequence number wraps back to the beginning), our malicious {\it leading packets} were implemented and the surgeon's packets were ignored due to the difference in the sequence numbers, and we effectively took control over the teleoperated procedure.

\noindent{\bf Force Reset:}
A interesting extension to the described {\it sequence number leading attack}, where we abused the way packet drops are handled and sequence numbers are processed, is the {\it force reset attack}, we were abused the robot's inherent safety mechanism, preventing the robot's arms from moving too fast or moving them outside of the allowed area. Every time the Raven's arms are commanded to move too fast, or go to an unsafe position, the robot's software imposes a system-wide halt, referred to as {\it software E-stop.} This is to protect both the electrical and mechanical components of the robot, as well as to ensure an extra level of safety for patients, and human operators standing near the robot. 


By sending a leading packet to the robot, where at least one of the changes in position or rotation is too large, and would cause the {\it Raven} to either go too fast or to go to a forbidden region, we are able to E-stop the robot.  Moreover, by repeatedly sending a malicious leading packet as the one just described, we are able to easily stop the robot from ever being properly reset, thus effectively making a surgical procedure impossible.
\vspace{-5pt}
\section{Mitigation Strategies}\label{sec:Mitigation_strategies}

\vspace{-2pt}
While the attacks that we demonstrated primarily target specific vulnerabilities of the \textit{Raven II}'s codebase, the identified exploits will have to be addressed for any teleoperated robotic system. There are several avenues for extended development to help secure these systems from the identified attacks. The first of these is maintaining a standard of communication robustness. The most important feature here is to provide a layer of security for all information passing between an operator and a teleoperated system. 

The injection attacks we demonstrated were successful due to the fact that valid packets were accepted by the robot from any source. For the {\it Raven II}, this was almost certainly a development oversight and is easy to fix. However, we need to consider the larger problem of how to protect against a more sophisticated packet spoofing attacks that also spoofs source IP and port information. One straightforward answer is to encrypt all data streams between the two endpoints rendering all but the man-in-the middle attacks impossible. An advantage that teleoperated systems have over many other network communication security problems is that there are likely dedicated staff at both ends of the system. This means that there exists an out-of-band communication method, such as texting or talking on the phone, to exchange a private piece of information that can be used to authenticate data streams. By {\it encrypting} and {\it authenticating} data streams between the surgeon's terminal and the robot, the ability of an attacker to initiate an attack that comprises an intention modification, manipulation, or hijack becomes severely hampered. 

In order to investigate the cost of encrypting all data exchanged between an operator and the {\it Raven}, that is, all data in the network, but not the side out-of-band communications, we used an intermediary computer with Intel Core2 Quad CPU processor running at 2.5GHz, to execute cryptographic tasks on. We acknowledge that the results obtained through this analysis do not necessarily represent the exact results we would have observed had we encrypted all packets closer to a surgeon and to the {\it Raven}, but for this analysis we only wanted to measure the added overhead of cryptographic operations. 
We used the Advanced Encryption Standard (AES) encryption method~\cite{daemen2002design}, and considered three different key lengths:128-bit, 192-bit and 256-bit. For all key lengths, no noticeable increase in CPU usage was observed, compared to the baseline case where the intermediary computer only received packet and forwarded them further. However, we observed an increase in memory usage, with the average increase of 3000KB. This increase value will likely be acceptable for the majority of teleoperated systems. As expected, we did not observe a significant memory usage difference between different key lengths. Thus the use of encryption and authentication has low cost and high benefits to telerobotic surgery, mitigating many analyzed attacks.
\section{Discussion}\label{sec:Discussion}

\subsection{Implications of Presented Attacks}\label{subsec:Implications}
Intention modification, manipulation and hijacking attacks pose not only technical challenges, but also considerable risks to patients, surgeons and robots. A compromised surgical robot in the midst of even a routine operation could potentially be used to inflict considerable internal wounds to a patient. 
Moreover, any extra procedure time, caused by a compromised system, may have severe consequences on a procedure outcome, as well as a patient's recovery. 
Finally, compromised data and video streams could pose a risk to patient privacy. A surgeon's actions,  haptic feedback and robot's video feed may all contain private and protected patient-related information. For instance, the images in the video stream may contain patient identifying features or may expose portions of the body that the patient would prefer to keep private. 

For surgeons, the possibility of surgery systems being compromised complicates the issue of legal responsibility for their actions during procedures. In intention modification or manipulation attacks, a surgeon does not have direct access to a robot and can only operate based on the exchanged information. In a compromised system, for example, haptic feedback may be modified to cause a surgeon to harm a patient. If one can claim that it was reasonable to expect that the surgeon should have noticed that haptic feedback was modified, than the resulting malpractice lawsuit might be strengthened.

Teleoperation security threats may have further implications for surgical robots themselves, since mounted attacks may cause robots to break, or to damage other nearby equipment in the operating room. Finally, any security holes in teleoperated systems present an existential threat to the field of surgical robotics as a whole. Even if attacks are rare, any harm caused by a surgical robot could undermine the public's faith in these systems. From a patient perspective, all the advantages in recovery or success rate that come from teleoperated surgery may not be worth the risk of having a potentially hijacked machine operate on them.

\vspace{-5pt}
\subsection{Challenges and Recommendations}
In addition to the stated implications, in these experiments we have encountered a few rather interesting challenges. We present those next, starting with the E-stop feature. 

{\bf E-stop feature:} In a benign case, an E-stop is a mechanism designed to improve safety of patients, near-by equipment and operators, and a robot. Our experiments have shown, however, that the existence of E-stop may actually lead to {\it decreased} safety and security of a robot and people in the  case of a compromised system. An attacker with sufficient knowledge of the system may easily abuse E-stop to render a robot unusable. For example, by occasionally sending {\it leading} packets, where at least one of the changes is sufficiently large, an attacker may cause the system to be permanently E-stopped. The challenge thus arises to reconcile the benefits of E-stop in the benign case with its possible negative consequences in the adversarial setting. 

{\bf Packets processing rate:} Another observed feature that can be turned into a security vulnerability is the fact that teleoperated robots (including {\it Raven II}) typically execute command packets as soon as they are received. This means that if a burst of packets is received, a robot may start moving very fast and in jerky motions. This may increase the wear on a robot's joints and motors, but more importantly, it may pose danger to a patient in surgery. And at the moment, an attacker can deliberately cause control commands to be received in bursts. {\it To protect against burst attacks, we propose limiting a robot's processing rate to a value sufficiently large to never be reached in benign scenarios, but low enough to protect the robot and the patient from harm due to a flood of commands.} 

{\bf Tension between real-time operation and security:} In order to ensure fast enough operation, many teleoperated systems resort to using datagram protocols. It is typically assumed that surgical tools' motions are continuous, and that transmission rates are sufficiently high, so that occasional benign packet losses have negligible effects on the overall procedure. Yet, in a hostile setting, an attacker with sufficient knowledge of the system may abuse the protocol, and specifically drop certain packets in order to cause maximal damage (harm), while being cautious about his own resources. Since datagram-based  protocols are likely to remain the preferred choice for teleoperated systems, an appropriate strategy to mitigate this type of threats will have to be found.

{\bf Tension between fast feedback and privacy:} Many of the attacks we presented may be mitigated by encrypting and authenticating {\it all} communication between a surgeon and a robot. Yet, due to sheer quantity of video data from the robot, and the real-time operation requirement, encrypting the entire feedback channel may not be feasible. In this case a trade-off between the real-time feedback and patient's privacy arises. {\it Based on the experimental results, however, we propose that authenticating all packets should be the minimum required feature for any teleoperated robotic system operating over a public network, so as to assure that packets from any other sources are never accepted as real.}

{\bf Security update to ITP:} We further consider the Interoperable telesurgery protocol (ITP). 
Some of the attacks we have mounted could have been easily prevented had the packets' sequence number processing been implemented differently (for example, sequence number leading attack), and had the protocol had checksum checking implemented (e.g., operator intent modification attack). {\it Sequence number processing and checksum checking are the minimal changes needed to increase security. Moreover, implementing these changes would not impact the interoperability.} 

{\bf Monitoring link and network status:} The presented hijacking attack relied on the fact that the {\it Raven II} ignores all out-of-order packets and only acts upon those packets that have a higher sequence number than the latest received packet. This essentially means that in addition to a surgeon's commands, an attacker can create one or more malicious packet streams and send those to a robot, and the robot will always execute packets with a higher sequence number. {\it In order to prevent these attacks, we propose that a mechanism to monitor link and network status should be used.} Such a mechanism should be able to notice that there are two or more streams of data, or that the number of out-or-order packets has increased, and should rise an alarm. A more advanced version of this mechanism could also prevent possible denial-of-service attacks that cannot be prevented though the use of the existing cryptographic methods. These attacks saturate network resources and cause packets to be delayed or dropped.
\vspace{-8pt}
\section{Conclusion}\label{sec:Conclusion}
In conclusion, the purpose of this paper is to increase awareness of security issues in cyber-physical systems. In the {\it Raven II}, we were able to breach several concerning elements of the system over a wide attack surface, and some extremely efficiently (with a single packet). Yet, some of these attacks could have easily been prevented by using well-established and readily-available security mechanisms, including encryption and authentication. Our experimental results show that incorporating these mechanisms into a telerobotic surgery system with an average-quality computer increases memory usage by only about 3000KB, while maintaining the system's real-time responsiveness. This increase is likely to be acceptable in telerobotic surgery.
 
We caution, however, that tensions between cyber security, safety and usability requirements of teleoperated robotic surgery  will render many existing security solutions infeasible for telerobotic surgery, thus requiring new security approaches to be developed. 
For example, encrypting and authenticating video feedback will likely cause an unacceptable decrease in packet throughput rate. 
Finally, we believe that presented concerns are not unique to teleoperated surgery, but are common to {\it all} teleoperated robots. Because of the wide variety of physical and digital capabilities these systems wield, telerobotic security needs to become front-and-center.

\small
\section*{ACKNOWLEDGMENT}
The authors thank UW Electrical Engineering undergraduate students, F. Ibrahim, L. Nguyen, and X. Ouyang for their help
with the experimental part of this project. We also gratefully acknowledge AppliedDexterity, and our colleagues, L. Cheng, D. Hu, K. Huang and A. Lewis, for their help with the {\it Raven II}.

\vspace{-5pt}
\scriptsize
\bibliographystyle{plain}
\bibliography{arXiv_April_2015}

\end{document}